\pdfoutput=1

\documentclass[11pt]{article}

\usepackage{EMNLP2023}

\usepackage{times}
\usepackage{latexsym}

\usepackage[T1]{fontenc}
\usepackage[utf8]{inputenc}

\usepackage{microtype}

\usepackage{inconsolata}

\usepackage{graphicx}
\usepackage{multirow}
\usepackage{makecell}
\usepackage{booktabs}
\usepackage{xcolor}
\usepackage{xspace}

\newcommand{\CRAC}{CRAC 2023 Shared Task\xspace}
\newcommand{\CRAClong}{CRAC 2023 Shared Task on Multilingual Coreference Resolution\xspace}

\newenvironment{citemize}{\begin{list}{$\bullet$}{\topsep=.1\smallskipamount\itemsep=0pt\parsep=1pt\labelwidth=.5em}}{\end{list}}

\title{ÚFAL~CorPipe at CRAC 2023: Larger Context Improves Multilingual~Coreference Resolution}

\author{Milan Straka \\
  Charles University, Faculty of Mathematics and Physics \\
  Institute of Formal and Applied Linguistics \\
  Malostranské nám. 25, Prague, Czech Republic \\
  \texttt{straka@ufal.mff.cuni.cz}
  }

\usepackage[absolute]{textpos}
\begin{document}
\begin{textblock}{16}(0,0.1)\centerline{This paper was published at \textbf{CRAC 2023} -- please cite the published version {\small\url{https://aclanthology.org/2023.crac-sharedtask.4/}}.}\end{textblock}
\maketitle
\begin{abstract}
  We present CorPipe, the winning entry to the \CRAClong. Our system is an
  improved version of our earlier multilingual coreference pipeline, and it
  surpasses other participants by a large margin of 4.5 percent points.
  CorPipe first performs mention detection, followed by coreference linking
  via an antecedent-maximization approach on the retrieved spans. Both
  tasks are trained jointly on all available corpora using a shared pretrained
  language model. Our main improvements comprise inputs larger than 512
  subwords and changing the mention decoding to support ensembling. The source
  code is available at {\small\url{https://github.com/ufal/crac2023-corpipe}}.
\end{abstract}

\section{Introduction}

The goal of coreference resolution is to identify and cluster multiple
occurrences of entities in the input text. The
\CRAClong~\cite{sharedtask-findings} aims to stimulate research in this area by
featuring coreference resolution on 17 corpora in 12 languages from the
CorefUD~1.1 dataset~\cite{CorefUD1.1_2023}. The current shared task is
a reiteration of the previous year's CRAC 2022 Shared
Task~\cite{zabokrtsky-etal-2022-findings}.

CorPipe, our entry to the \CRAC, is an improved version of our earlier
multilingual coreference pipeline~\cite{straka-strakova-2022-ufal}, which was
the winner of the last year's shared task. Our system first performs mention
detection, followed by the coreference linking via an antecedent-maximization
approach on the retrieved spans. However, CorPipe is not a pure pipeline,
because we train both tasks jointly using a shared pretrained language model.
Performing mention detection first avoids the challenge of end-to-end systems
that need to consider an overwhelming number of possible spans, and also permits
recognition of single-mention entities. Finally, all our models are
multilingual and are trained on all available corpora.

Our contributions are as follows:
\begin{citemize}
  \item We present a winning entry to the \CRAC with state-of-the-art results,
    surpassing other shared task participants by a large margin of 4.5 percent
    points.
  \item We improve our last year's system by (a) increasing the size of the
    inputs during prediction, while keeping it smaller during training, (b)
    using larger pretrained language models, (c) proposing a different mention
    decoding approach, that allows (d) implementing ensembling to further
    improve the performance.
  \item We perform a thorough examination of the newly introduced components.
  \item The source code of our system is available at {\small\url{https://github.com/ufal/crac2023-corpipe}}.
\end{citemize}

\section{Related Work}

While coreference resolution was traditionally carried out by first performing
mention detection followed by coreference linking (clustering), recent
approaches are often end-to-end~\cite{lee-etal-2017-end,lee-etal-2018-higher}.
Likewise, the baseline of CRAC 2022 and 2023 Shared
Tasks~\cite{prazak-etal-2021-multilingual} as well as the CRAC 2022 second-best
solution~\cite{prazak-konopik-2022-end} follow this approach.

\looseness1
The recent work of \citet{bohnet-etal-2023-coreference} pushes the
end-to-end approach even further, solving both mention detection and
coreference linking jointly via a text-to-text paradigm, reaching
state-of-the-art results on the CoNLL 2012 dataset~\cite{pradhan-etal-2012-conll}.
Given that our system uses the same pretrained encoder but a custom
decoder designed specifically for coreference resolution instead
of a general but pretrained decoder, it would be interesting to perform a direct
comparison of these systems.

\begin{figure*}[t!]
    \centering
    \includegraphics[width=.905\hsize]{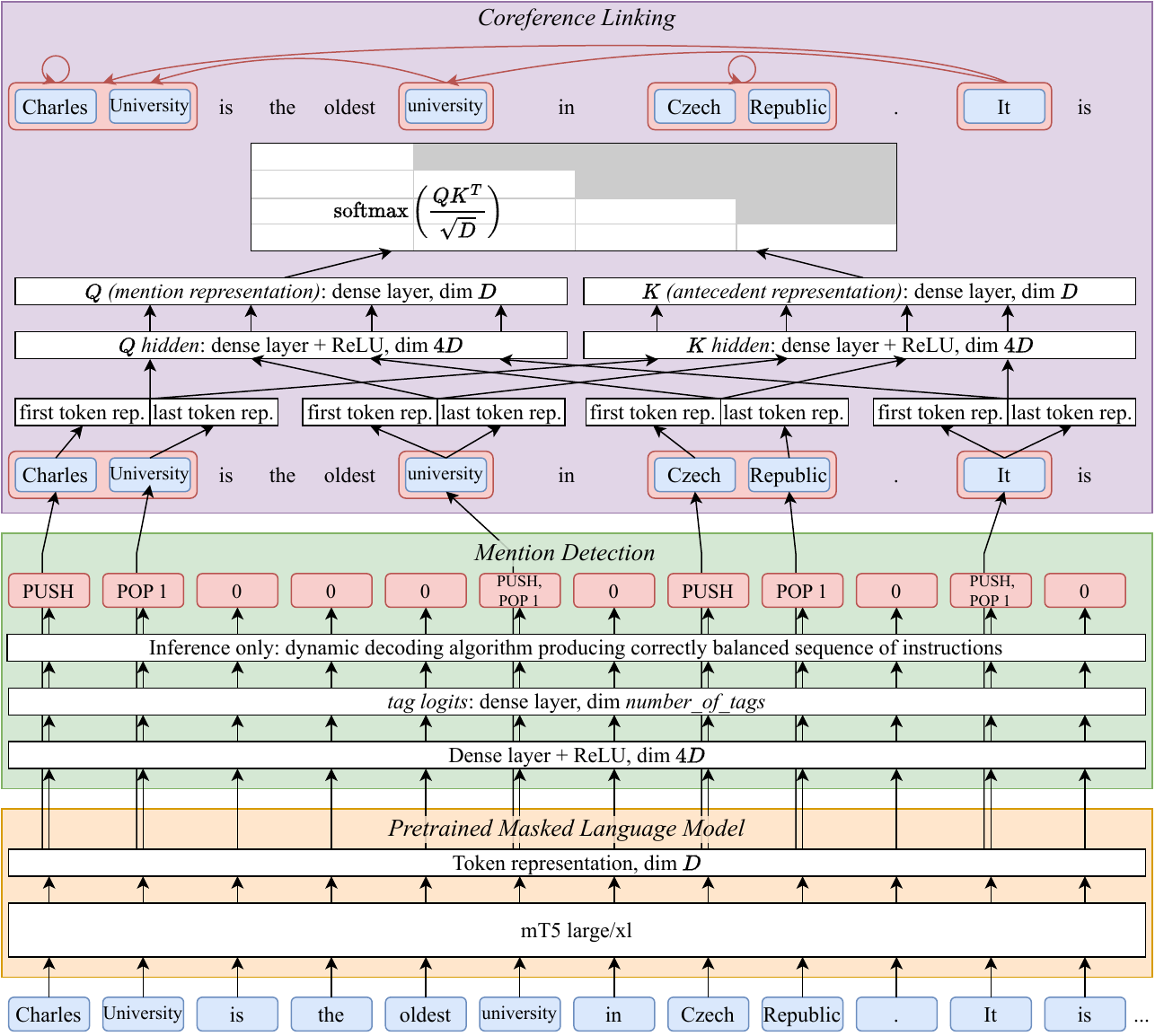}
    \caption{The proposed CorPipe model architecture.}
    \label{fig:architecture}
\end{figure*}

\section{CorPipe Architecture}

The CorPipe architecture is based heavily on our earlier
system~\cite{straka-strakova-2022-ufal}, which won the CRAC 2022 Shared
Task~\cite{zabokrtsky-etal-2022-findings}. We describe just the changes we
propose; please refer to~\cite{straka-strakova-2022-ufal} for the description
of our original system.

In short, our system first obtains a contextualized representation of the input
by employing a pretrained model. These representations are then used first to
perform mention detection, and then, together with the predicted mentions, to
perform coreference linking. The mentions are predicted one sentence at a time,
but both previous and following contexts are included up to the specified
\textit{context length}. The architecture overview is displayed in
Figure~\ref{fig:architecture}.

\subsection{The mT5 Pretrained Models}

In the original architecture, we employed large-sized models
XLM-R large~\citep{conneau-etal-2020-unsupervised}
and RemBERT~\citep{chung2021rethinking}. However, even bigger
models consistently deliver better performance
in various applications
\citep{kale-rastogi-2020-text,xue-etal-2021-mt5,rothe-etal-2021-simple,bohnet-etal-2023-coreference}.
We therefore decided to utilize the largest possible pretrained multilingual
model. To our best knowledge, we are aware of a single family of such models,
the mT5~\cite{xue-etal-2021-mt5}, a multilingual variant of
the encoder-decoder pretrained model T5~\cite{kale-rastogi-2020-text}
based on the Transformer architecture~\cite{vaswani-etal-2017-attention}.%
\footnote{The ByT5~\cite{xue-etal-2022-byt5}, a byte version of multilingual T5, is also
available, but because it represents words as individual UTF-8 bytes, it
processes smaller inputs compared to mT5, which is undesirable for
coreference resolution.}

The mT5 pretrained models have one more considerable advantage -- because of
relative positional embeddings, they are capable of processing inputs longer
than 512 subwords, compared to both XLM-R large and RemBERT. In
Section~\ref{sec:plm_and_context}, we demonstrate that processing longer inputs
is advantageous for coreference resolution.

\subsection{Mention Decoding}

In the original architecture, we reduce the representation of embedded and
possibly crossing mentions to a sequence classification problem using an
extension of BIO encoding. Each input token is assigned a single tag, which
is a concatenation of a sequence of stack-manipulating instructions:
\begin{citemize}
  \item any number of \texttt{POP}$(i)$ instructions, each closing an opened mention
    from the stack. To support crossing mentions, any mention on the stack (not
    just the top one) can be closed, identified by its index $i$ from the top of
    the stack (i.e., \texttt{POP}$(1)$ closes the mention on the top of the
    stack, \texttt{POP}$(2)$ closes the mention below the top of the stack);
  \item any number of \texttt{PUSH} instructions, each starting a new mention
    added to the top of the stack;
  \item any number of \texttt{POP}$(1)$ instructions, each closing a single-token
    mention started by a \texttt{PUSH} instruction from the same tag (such
    single-token mentions could be also represented by a dedicated instruction
    like \texttt{UNIT}, but we prefer smaller number of instructions).
\end{citemize}
\noindent To produce hopefully valid (well-balanced) sequences of tags, we
originally used a linear-chain conditional random fields (CRF; \citealt{lafferty-etal-2021-conditional}).
Because of the Markovian property, every tag had to be parametrized also
with the size of the stack before the first instruction (we call these
tags the \textit{depth-dependent tags}).

The described approach has two drawbacks. First, the predicted sequence
of tags might still be unbalanced (which we observed repeatedly in the
predictions). Furthermore, it would be more challenging to perform ensembling,
because every model would have a different sequence-based partition
function.\footnote{When ensembling models, we average the
\textit{distributions} the models predict; in other words, unnormalized logits
must first be normalized into (log-)probabilities. While this is
straightforward for simple classification, CRF models normalize over all
possible label sequences. Ensembling several CRF models would therefore require
that, during each step of the sequential decoding of token labels, every model
computed the \textit{(log-)probabilities} of all sequences with the label in
question conditioned on the already decoded labels. Such an algorithm would
have the same asymptotic complexity as the usual CRF decoding times the number
of models. However, we did not implement it ourselves.}

To alleviate both mentioned issues, we propose to replace the CRF with
per-token classification during training and perform a constrained
dynamic programming decoding during inference using the Viterbi
algorithm.\footnote{The decoding algorithm differs from CRF decoding in just
two aspects: (a) the logits are normalized into log-probabilities for each
token separately, (b) the transition matrix only forbids invalid transitions,
all valid transitions have the same weight.}
Such approach admits ensembling in a straightforward manner by averaging
predicted distributions for each token independently.

Without the CRF, the tags no longer need to be parametrized by the current
size of the stack -- the depth of the stack can be tracked just during decoding (we
consider stack depths of at most 10; Section~\ref{sec:decoding_ablations}
demonstrates that depth 3 is actually sufficient). Such
\textit{depth-independent tags} have the advantage of being
scarcer,\footnote{There are 54 and 207 unique \textit{depth-independent} and
\textit{depth-dependent tags} in the whole
training data, respectively.}
admitting better statistical efficiency, and we utilize them in our primary
submission. The comparison of both tag sets
as well as the CRF and dynamic programmic decoding is performed in
Section~\ref{sec:decoding_ablations}.

\begin{table*}[t]
  \centering\setlength{\tabcolsep}{3.3pt}\scriptsize
  \renewcommand\cellset{\renewcommand\arraystretch{0.85}}
    \catcode`@ = 13\def@{\bfseries}
  \begin{tabular}{lcccccccccccccccccc}
    \toprule
      System & Avg &
      \texttt{ca} &
      \makecell[c]{\texttt{cs} \\ \texttt{\kern-.2em pcedt\kern-.2em}} &
      \makecell[c]{\texttt{cs} \\ \texttt{pdt}} &
      \makecell[c]{\texttt{de} \\ \texttt{parc}} &
      \makecell[c]{\texttt{de} \\ \texttt{pots}} &
      \makecell[c]{\texttt{en} \\ \texttt{gum}} &
      \makecell[c]{\texttt{en} \\ \texttt{parc}} &
      \texttt{es} &
      \texttt{fr} &
      \makecell[c]{\texttt{hu} \\ \texttt{\kern-.2em korko\kern-.2em}} &
      \makecell[c]{\texttt{hu} \\ \texttt{\kern-.2em szege\kern-.2em}} &
      \texttt{lt} &
      \makecell[c]{\texttt{no} \\ \texttt{\kern-.2em bookm\kern-.2em}} &
      \makecell[c]{\texttt{no} \\ \texttt{\kern-.2em nynor\kern-.2em}} &
      \texttt{pl} &
      \texttt{ru} &
      \texttt{tr} \\
  \midrule
    \textbf{ÚFAL CorPipe} & @\makecell[c]{74.90 \\ 1} & @\makecell[c]{82.59 \\ 1} & @\makecell[c]{79.33 \\ 1} & @\makecell[c]{79.20 \\ 1} & @\makecell[c]{72.12 \\ 1} & @\makecell[c]{71.09 \\ 1} & @\makecell[c]{76.57 \\ 1} & @\makecell[c]{69.86 \\ 1} & @\makecell[c]{83.39 \\ 1} & @\makecell[c]{69.82 \\ 1} & @\makecell[c]{68.92 \\ 1} & @\makecell[c]{69.47 \\ 1} & @\makecell[c]{75.87 \\ 1} & @\makecell[c]{78.74 \\ 1} & @\makecell[c]{78.77 \\ 1} & @\makecell[c]{79.54 \\ 1} & @\makecell[c]{82.46 \\ 1} & @\makecell[c]{55.63 \\ 1} \\
    Anonymous & \makecell[c]{70.41 \\ 2} & \makecell[c]{79.51 \\ 2} & \makecell[c]{75.88 \\ 2} & \makecell[c]{76.39 \\ 2} & \makecell[c]{64.37 \\ 3} & \makecell[c]{68.24 \\ 5} & \makecell[c]{72.29 \\ 2} & \makecell[c]{59.02 \\ 3} & \makecell[c]{80.52 \\ 2} & \makecell[c]{66.13 \\ 2} & \makecell[c]{64.65 \\ 3} & \makecell[c]{66.25 \\ 2} & \makecell[c]{70.09 \\ 2} & \makecell[c]{75.32 \\ 2} & \makecell[c]{73.33 \\ 2} & \makecell[c]{77.58 \\ 2} & \makecell[c]{80.19 \\ 2} & \makecell[c]{47.22 \\ 2} \\
    Ondfa & \makecell[c]{69.19 \\ 3} & \makecell[c]{76.02 \\ 3} & \makecell[c]{74.82 \\ 3} & \makecell[c]{74.67 \\ 3} & \makecell[c]{71.86 \\ 2} & \makecell[c]{69.37 \\ 3} & \makecell[c]{71.56 \\ 3} & \makecell[c]{61.62 \\ 2} & \makecell[c]{77.18 \\ 3} & \makecell[c]{60.32 \\ 4} & \makecell[c]{66.38 \\ 2} & \makecell[c]{65.75 \\ 4} & \makecell[c]{68.52 \\ 3} & \makecell[c]{72.39 \\ 4} & \makecell[c]{70.91 \\ 4} & \makecell[c]{76.90 \\ 3} & \makecell[c]{76.50 \\ 4} & \makecell[c]{41.52 \\ 4} \\
    McGill & \makecell[c]{65.43 \\ 4} & \makecell[c]{71.75 \\ 4} & \makecell[c]{67.67 \\ 7} & \makecell[c]{70.88 \\ 4} & \makecell[c]{41.58 \\ 7} & \makecell[c]{70.20 \\ 2} & \makecell[c]{66.72 \\ 4} & \makecell[c]{47.27 \\ 4} & \makecell[c]{73.78 \\ 4} & \makecell[c]{65.17 \\ 3} & \makecell[c]{60.74 \\ 4} & \makecell[c]{65.93 \\ 3} & \makecell[c]{65.77 \\ 6} & \makecell[c]{73.73 \\ 3} & \makecell[c]{72.43 \\ 3} & \makecell[c]{76.14 \\ 4} & \makecell[c]{77.28 \\ 3} & \makecell[c]{45.28 \\ 3} \\
    DeepBlueAI & \makecell[c]{62.29 \\ 5} & \makecell[c]{67.55 \\ 7} & \makecell[c]{70.38 \\ 4} & \makecell[c]{69.93 \\ 5} & \makecell[c]{48.81 \\ 5} & \makecell[c]{63.90 \\ 7} & \makecell[c]{63.58 \\ 6} & \makecell[c]{43.33 \\ 5} & \makecell[c]{69.52 \\ 5} & \makecell[c]{55.69 \\ 6} & \makecell[c]{54.38 \\ 5} & \makecell[c]{63.14 \\ 5} & \makecell[c]{66.75 \\ 4} & \makecell[c]{69.86 \\ 6} & \makecell[c]{68.53 \\ 5} & \makecell[c]{73.11 \\ 5} & \makecell[c]{74.41 \\ 5} & \makecell[c]{36.14 \\ 8} \\
    DFKI-Adapt & \makecell[c]{61.86 \\ 6} & \makecell[c]{68.21 \\ 6} & \makecell[c]{68.72 \\ 5} & \makecell[c]{67.34 \\ 6} & \makecell[c]{52.52 \\ 4} & \makecell[c]{69.28 \\ 4} & \makecell[c]{65.11 \\ 5} & \makecell[c]{36.87 \\ 7} & \makecell[c]{69.19 \\ 6} & \makecell[c]{58.96 \\ 5} & \makecell[c]{51.53 \\ 7} & \makecell[c]{58.56 \\ 6} & \makecell[c]{66.01 \\ 5} & \makecell[c]{70.05 \\ 5} & \makecell[c]{68.21 \\ 6} & \makecell[c]{67.98 \\ 6} & \makecell[c]{72.48 \\ 6} & \makecell[c]{40.67 \\ 5} \\
    Morfbase & \makecell[c]{59.53 \\ 7} & \makecell[c]{68.23 \\ 5} & \makecell[c]{64.89 \\ 8} & \makecell[c]{64.74 \\ 8} & \makecell[c]{39.96 \\ 9} & \makecell[c]{64.87 \\ 6} & \makecell[c]{62.80 \\ 8} & \makecell[c]{40.81 \\ 6} & \makecell[c]{69.01 \\ 7} & \makecell[c]{53.18 \\ 8} & \makecell[c]{52.91 \\ 6} & \makecell[c]{56.41 \\ 7} & \makecell[c]{64.08 \\ 7} & \makecell[c]{68.17 \\ 7} & \makecell[c]{66.35 \\ 7} & \makecell[c]{67.88 \\ 7} & \makecell[c]{68.53 \\ 8} & \makecell[c]{39.22 \\ 6} \\
    BASELINE$^\dagger$ & \makecell[c]{56.96 \\ 8} & \makecell[c]{65.26 \\ 8} & \makecell[c]{67.72 \\ 6} & \makecell[c]{65.22 \\ 7} & \makecell[c]{44.11 \\ 6} & \makecell[c]{57.13 \\ 9} & \makecell[c]{63.08 \\ 7} & \makecell[c]{35.19 \\ 8} & \makecell[c]{66.93 \\ 8} & \makecell[c]{55.31 \\ 7} & \makecell[c]{40.71 \\ 9} & \makecell[c]{55.32 \\ 8} & \makecell[c]{63.57 \\ 8} & \makecell[c]{65.10 \\ 9} & \makecell[c]{65.78 \\ 8} & \makecell[c]{66.08 \\ 8} & \makecell[c]{69.03 \\ 7} & \makecell[c]{22.75 \\ 9} \\
    DFKI-MPrompt & \makecell[c]{53.76 \\ 9} & \makecell[c]{55.45 \\ 9} & \makecell[c]{60.39 \\ 9} & \makecell[c]{56.13 \\ 9} & \makecell[c]{40.34 \\ 8} & \makecell[c]{59.75 \\ 8} & \makecell[c]{57.83 \\ 9} & \makecell[c]{34.32 \\ 9} & \makecell[c]{58.31 \\ 9} & \makecell[c]{52.96 \\ 9} & \makecell[c]{44.53 \\ 8} & \makecell[c]{48.79 \\ 9} & \makecell[c]{56.52 \\ 9} & \makecell[c]{65.12 \\ 8} & \makecell[c]{62.99 \\ 9} & \makecell[c]{61.15 \\ 9} & \makecell[c]{61.96 \\ 9} & \makecell[c]{37.44 \\ 7} \\
  \bottomrule
\end{tabular}

  \caption{Official results of \CRAC~on the test set (CoNLL score in \%). The system $^\dagger$ is described in \citet{prazak-etal-2021-multilingual}; the rest in \citet{sharedtask-findings}.}
  \label{tab:official_treebanks}
\end{table*}

\subsection{Multilingual Training Data}
\label{sec:multilingual_training_data}

All our models are trained on all 17 CorefUD 1.1 corpora. Given that
their size range from tiny (457 training sentences in \texttt{de}
and \texttt{en parcorfull}) to large (almost 40k training sentences
in \texttt{cs pdt} and \texttt{cs pcedt}), we try to level the
individual corpora performances by sub-/over-sampling the datasets.
Concretely, we sample each batch example (a sentence with its context)
proportionally to \textit{mix ratios}, the corpora-specific weights. We
consider the following possibilities:

\begin{citemize}
  \item \textit{uniform}: we sample uniformly from all corpora, ignoring
    their sizes;
  \item \textit{linear}: we sample proportionally to the sizes of individual
    corpora;
  \item \textit{square root}: following \cite{van-der-goot-etal-2021-massive},
    we sample proportionally to the square roots of corpora sizes;
  \item \textit{logarithmic}: similar to \cite{straka-strakova-2022-ufal},
    we sample proportionally to the corpora sizes logarithms, which are linearly
    rescaled so that the largest corpus is ten times more probable than the
    smallest corpus.
\end{citemize}

Since different corpora might require particular annotations, we also
consider adding a \textit{corpus id} subword (dataset label) to the input to
indicate the dataset of origin and the required style of annotations.
These \textit{corpus ids}, evaluated already in
\cite{straka-strakova-2022-ufal}, are just a different implementation of
treebank embeddings proposed in \citet{stymne-etal-2018-parser}.

Our primary submission relies on \textit{logarithmic} mix ratios with
\textit{corpus ids}. The concrete values of all proposed mix ratios together
with their performance comparison are presented in
Section~\ref{sec:mix_ratios}.

\subsection{Training}
\label{sec:training}

\looseness-1
When utilizing the mT5 pretrained models, we train CorPipe models with the
Adafactor optimizer~\cite{adafactor} using a slanted triangular
learning schedule -- we first linearly increase the learning rate from
0 to \texttt{5e-4} in the first 10\% of the training, and then linearly
decay it to 0 at the end of the training. The models are trained for
15 epochs, each comprising 8000 batches. For models up to size large,
we utilize batch size 8, which is the maximum one fitting on a single A100
GPU with 40GB RAM. The xl-sized models are trained on four 40GB A100,
with a maximum possible batch size 12. The training took 10 and 20 hours for
the mT5-large and mT5-xl models, respectively.

For the XLM-R and RemBERT ablation experiments, we utilize the lazy variant of
the Adam optimizer~\cite{kingma-and-ba-2015} and the learning rates of
\texttt{2e-5} and \texttt{1e-5}, respectively.

All classification heads employ label
smoothing~\citep{szegedy-etal-2015-rethinking} of 0.2.

During training, we use \textit{context length} of 512 subwords and limit the
right context length to 50, but we use \textit{context length} of 2560
subwords during inference with the mT5 models.

The competition submissions were selected from a pool of 30 models based on
mT5-large and mT5-xl pretrained models with different random seeds and slightly
perturbed hyperparameters,\footnote{Learning rate \texttt{5e-4}, \texttt{6e-4},
\texttt{7e-4}; double or quadruple batch size; 8k or 10k batches per
epoch.} by considering for each corpus the best performing checkpoint of every
epoch of every trained model. Our primary submission is for each corpus an
ensemble of 3 best checkpoints of 3 models.\footnote{We implemented ensembling
by loading each model to its dedicated A100 GPU, thus parallelizing the
execution of the individual models.}

\begin{table}[t]
  \centering\setlength{\tabcolsep}{3.1pt}\scriptsize
    \catcode`@ = 13\def@{\bfseries}
  \begin{tabular}{lcccc}
    \toprule
      System & Head-match & Partial-match & Exact-match & +Singletons \\
  \midrule
    \textbf{ÚFAL CorPipe} & @74.90 (1) & @73.33 (1) & @71.46 (1) & @76.82 (1) \\
    Anonymous & 70.41 (2) & 69.23 (2) & 67.09 (2) & 73.20 (2) \\
    Ondfa & 69.19 (3) & 68.93 (3) & 53.01 (8) & 68.37 (3) \\
    McGill & 65.43 (4) & 64.56 (4) & 63.13 (3) & 68.23 (4) \\
    DeepBlueAI & 62.29 (5) & 61.32 (5) & 59.95 (4) & 54.51 (5) \\
    DFKI-Adapt & 61.86 (6) & 60.83 (6) & 59.18 (5) & 53.94 (6) \\
    Morfbase & 59.53 (7) & 58.49 (7) & 56.89 (6) & 52.07 (7) \\
    BASELINE & 56.96 (8) & 56.28 (8) & 54.75 (7) & 49.32 (8) \\
    DFKI-MPrompt & 53.76 (9) & 51.62 (9) & 50.42 (9) & 46.83 (9) \\
  \bottomrule
\end{tabular}

  \caption{Official results of \CRAC~on the test set with various metrics in \%.}
  \label{tab:official_metrics}
\end{table}

\begin{table*}[t]
  \centering\setlength{\tabcolsep}{2.85pt}\scriptsize
  \renewcommand\arraystretch{0.84}
      \catcode`@ = 13\def@{\bfseries}
    \catcode`! = 13\def!{\itshape}
    \begin{tabular}{lrrrrrrrrrrrrrrrrrr}
      \toprule
        Submission & Avg &
        \texttt{ca} &
        \makecell[c]{\texttt{cs} \\ \texttt{\kern-.2em pcedt\kern-.2em}} &
        \makecell[c]{\texttt{cs} \\ \texttt{pdt}} &
        \makecell[c]{\texttt{de} \\ \texttt{parc}} &
        \makecell[c]{\texttt{de} \\ \texttt{pots}} &
        \makecell[c]{\texttt{en} \\ \texttt{gum}} &
        \makecell[c]{\texttt{en} \\ \texttt{parc}} &
        \texttt{es} &
        \texttt{fr} &
        \makecell[c]{\texttt{hu} \\ \texttt{\kern-.2em korko\kern-.2em}} &
        \makecell[c]{\texttt{hu} \\ \texttt{\kern-.2em szege\kern-.2em}} &
        \texttt{lt} &
        \makecell[c]{\texttt{no} \\ \texttt{\kern-.2em bookm\kern-.2em}} &
        \makecell[c]{\texttt{no} \\ \texttt{\kern-.2em nynor\kern-.2em}} &
        \texttt{pl} &
        \texttt{ru} &
        \texttt{tr} \\
    \midrule
Original CorPipe 2022 & \textcolor{black}{70.3} & \textcolor{black}{79.9} & \textcolor{black}{76.0} & \textcolor{black}{76.8} & \textcolor{black}{63.3} & \textcolor{black}{@72.6} & \textcolor{black}{72.3} & \textcolor{black}{57.6} & \textcolor{black}{81.2} & \textcolor{black}{65.4} & \textcolor{black}{66.2} & \textcolor{black}{65.4} & \textcolor{black}{68.6} & \textcolor{black}{75.4} & \textcolor{black}{73.6} & \textcolor{black}{79.0} & \textcolor{black}{78.4} & \textcolor{black}{42.5} \\
Single mT5 large model & \textcolor{blue!56.2!black}{+2.6} & \textcolor{blue!80.1!black}{+2.2} & \textcolor{blue!63.1!black}{+2.1} & \textcolor{blue!32.8!black}{+0.8} & \textcolor{blue!51.0!black}{+6.7} & \textcolor{red!29.3!black}{--\kern 0.04em 1.2} & \textcolor{blue!37.7!black}{+1.6} & \textcolor{blue!32.4!black}{+4.0} & \textcolor{blue!40.7!black}{+0.9} & \textcolor{blue!0.9!black}{+0.1} & \textcolor{blue!60.9!black}{+1.6} & \textcolor{blue!80.4!black}{+3.3} & \textcolor{blue!100.0!black}{@+7.4} & \textcolor{blue!100.0!black}{@+3.5} & \textcolor{blue!41.3!black}{+2.2} & \textcolor{red!66.3!black}{--\kern 0.04em 0.5} & \textcolor{blue!58.1!black}{+2.4} & \textcolor{blue!58.2!black}{+7.6} \\
Single mT5 xl model & \textcolor{blue!59.2!black}{+2.7} & \textcolor{blue!72.6!black}{+2.0} & \textcolor{blue!60.7!black}{+2.0} & \textcolor{blue!64.3!black}{+1.5} & \textcolor{blue!20.7!black}{+2.7} & \textcolor{red!72.7!black}{--\kern 0.04em 3.0} & \textcolor{blue!66.5!black}{+2.9} & \textcolor{blue!55.9!black}{+6.8} & \textcolor{blue!71.5!black}{+1.6} & \textcolor{blue!59.1!black}{+2.6} & \textcolor{red!100.0!black}{--\kern 0.04em 0.7} & \textcolor{blue!100.0!black}{@+4.1} & \textcolor{blue!64.0!black}{+4.7} & \textcolor{blue!95.1!black}{+3.3} & \textcolor{blue!71.0!black}{+3.7} & \textcolor{red!41.0!black}{--\kern 0.04em 0.3} & \textcolor{blue!64.3!black}{+2.6} & \textcolor{blue!78.7!black}{+10.3} \\
Per-treebank best mT5 model & \textcolor{blue!74.6!black}{+3.4} & \textcolor{blue!94.7!black}{+2.6} & \textcolor{blue!52.0!black}{+1.7} & \textcolor{blue!68.4!black}{+1.6} & \textcolor{blue!100.0!black}{@+13.1} & \textcolor{red!100.0!black}{--\kern 0.04em 4.1} & \textcolor{blue!75.5!black}{+3.2} & \textcolor{blue!84.3!black}{+10.3} & \textcolor{blue!53.4!black}{+1.2} & \textcolor{blue!74.1!black}{+3.3} & \textcolor{red!26.6!black}{--\kern 0.04em 0.2} & \textcolor{blue!48.3!black}{+2.0} & \textcolor{blue!89.3!black}{+6.6} & \textcolor{blue!84.7!black}{+3.0} & \textcolor{blue!80.7!black}{+4.2} & \textcolor{red!100.0!black}{--\kern 0.04em 0.8} & \textcolor{blue!94.3!black}{+3.8} & \textcolor{blue!58.2!black}{+7.6} \\
@Per-treebank 3-model ensemble & \textcolor{blue!100.0!black}{@+4.6} & \textcolor{blue!100.0!black}{@+2.7} & \textcolor{blue!100.0!black}{@+3.3} & \textcolor{blue!100.0!black}{@+2.4} & \textcolor{blue!67.3!black}{+8.8} & \textcolor{red!36.9!black}{--\kern 0.04em 1.5} & \textcolor{blue!100.0!black}{@+4.3} & \textcolor{blue!100.0!black}{@+12.3} & \textcolor{blue!100.0!black}{@+2.2} & \textcolor{blue!100.0!black}{@+4.4} & \textcolor{blue!100.0!black}{@+2.7} & \textcolor{blue!99.8!black}{+4.1} & \textcolor{blue!98.1!black}{+7.3} & \textcolor{blue!95.7!black}{+3.3} & \textcolor{blue!100.0!black}{@+5.2} & \textcolor{blue!100.0!black}{@+0.5} & \textcolor{blue!100.0!black}{@+4.1} & \textcolor{blue!100.0!black}{@+13.1} \\
    \midrule
!Per-treebank 8-model ensemble & !\textcolor{blue!100.0!black}{@+4.9} & !\textcolor{blue!100.0!black}{@+3.3} & !\textcolor{blue!100.0!black}{@+3.3} & !\textcolor{blue!100.0!black}{@+2.7} & !\textcolor{blue!58.7!black}{+7.7} & !\textcolor{red!19.2!black}{--\kern 0.04em 0.8} & !\textcolor{blue!97.2!black}{+4.2} & !\textcolor{blue!100.0!black}{@+13.4} & !\textcolor{blue!100.0!black}{@+2.3} & !\textcolor{blue!73.0!black}{+3.2} & !\textcolor{blue!100.0!black}{@+3.3} & !\textcolor{blue!100.0!black}{@+5.4} & !\textcolor{blue!100.0!black}{@+7.8} & !\textcolor{blue!100.0!black}{@+4.2} & !\textcolor{blue!100.0!black}{@+5.4} & !\textcolor{blue!100.0!black}{@+0.8} & !\textcolor{blue!100.0!black}{@+4.2} & !\textcolor{blue!100.0!black}{@+14.0} \\
    \bottomrule
    \end{tabular}

  \caption{Official results of ablation experiments on the test set (CoNLL score in \%).
  The 8-model ensemble (in italics) was evaluated during the post-competition phase.}
  \label{tab:test_results}
\end{table*}

\section{Shared Task Results}
\label{sec:results}

The official results of the \CRAC are presented in Table~\ref{tab:official_treebanks}.
Our CorPipe system delivers the best overall score of 74.9\%, surpassing the other
participants by a large margin of 4.5 percent points, and also achieves the
best scores for all individual corpora.

\subsection{Results of Additional Metrics}

The \CRAC primary metric employs \textit{head matching}, where a predicted
mention is considered correct if it has the same mention head as the gold
mention, and excludes \textit{singletons}. Comparison with other metrics is
performed in Table~\ref{tab:official_metrics}. Apart from the head matching,
the organizers evaluated also \textit{partial matching} (a predicted mention is
correct if it is a subsequence of the gold mention and contains the gold
mention head), \textit{exact matching} (a predicted mention is correct if it is
exactly equal to the gold mention), and head matching including
\textit{singletons} (entities with a single mention).

The ranking of all systems is unchanged in all evaluated metrics, with a single
exception -- the system \textit{Ondfa} exhibits low exact-matching performance,
presumably because it reduces predicted mentions to just their heads.%
\footnote{ Reducing mentions to heads was a strategy for improving partial-matching
score in the previous edition of the shared task; with the head-matching score,
it can be avoided, which allows also correct evaluation of the exact matching.}

\begin{table*}[t]
  \centering\setlength{\tabcolsep}{2.95pt}\scriptsize
  \renewcommand\arraystretch{1.02}
\catcode`@ = 13\def@{\bfseries}
\catcode`! = 13\def!{\itshape}
\begin{tabular}{lrrrrrrrrrrrrrrrrrr}
  \toprule
    Configuration & Avg &
    \texttt{ca} &
    \makecell[c]{\texttt{cs} \\ \texttt{\kern-.2em pcedt\kern-.2em}} &
    \makecell[c]{\texttt{cs} \\ \texttt{pdt}} &
    \makecell[c]{\texttt{de} \\ \texttt{parc}} &
    \makecell[c]{\texttt{de} \\ \texttt{pots}} &
    \makecell[c]{\texttt{en} \\ \texttt{gum}} &
    \makecell[c]{\texttt{en} \\ \texttt{parc}} &
    \texttt{es} &
    \texttt{fr} &
    \makecell[c]{\texttt{hu} \\ \texttt{\kern-.2em korko\kern-.2em}} &
    \makecell[c]{\texttt{hu} \\ \texttt{\kern-.2em szege\kern-.2em}} &
    \texttt{lt} &
    \makecell[c]{\texttt{no} \\ \texttt{\kern-.2em bookm\kern-.2em}} &
    \makecell[c]{\texttt{no} \\ \texttt{\kern-.2em nynor\kern-.2em}} &
    \texttt{pl} &
    \texttt{ru} &
    \texttt{tr} \\
  \midrule
  \noalign{\vskip 1pt}\multicolumn{19}{l}{\textsc{A) Context Sizes for the mT5-large Model}} \\[3pt]
    mT5-large 512 & 72.8 & 78.1 & 78.1 & 76.9 & @70.7 & @75.4 & 75.6 & @67.4 & 80.3 & 68.6 & @70.6 & 67.3 & 77.4 & 77.8 & 78.7 & 75.8 & 71.1 & 48.6 \\
    mT5-large 256 & \textcolor{red!100.0!black}{--\kern 0.04em 5.9} & \textcolor{red!100.0!black}{--\kern 0.04em 8.8} & \textcolor{red!100.0!black}{--\kern 0.04em 4.0} & \textcolor{red!100.0!black}{--\kern 0.04em 5.3} & \textcolor{red!100.0!black}{--\kern 0.04em 7.1} & \textcolor{red!100.0!black}{--\kern 0.04em 3.2} & \textcolor{red!100.0!black}{--\kern 0.04em 5.3} & \textcolor{red!100.0!black}{--\kern 0.04em 11.7} & \textcolor{red!100.0!black}{--\kern 0.04em 6.0} & \textcolor{red!100.0!black}{--\kern 0.04em 4.1} & \textcolor{red!100.0!black}{--\kern 0.04em 2.9} & \textcolor{red!100.0!black}{--\kern 0.04em 4.5} & \textcolor{red!100.0!black}{--\kern 0.04em 8.6} & \textcolor{red!100.0!black}{--\kern 0.04em 6.4} & \textcolor{red!100.0!black}{--\kern 0.04em 6.4} & \textcolor{red!100.0!black}{--\kern 0.04em 4.8} & \textcolor{red!100.0!black}{--\kern 0.04em 6.7} & \textcolor{red!100.0!black}{--\kern 0.04em 4.6} \\
\bottomrule
\end{tabular}

  \caption{Ablation experiments evaluated on the development sets (CoNLL score in \%). We report the average of best 5 out of 7 runs, using for every corpus the single epoch achieving the highest average 5-run score. The runs in italics use largest context length not exceeding 512 subwords when tokenized with the mT5 tokenizer.}
  \label{tab:models_ablations}
\end{table*}

\subsection{Results of Our Additional Submissions}

To quantify this year's CorPipe improvements, we present the official results
of our additional submissions in Table~\ref{tab:test_results}.

We first trained the original CorPipe on this year's data, achieving a 70.3\%
CoNLL score, which is 0.1 percent points below the second-best submission.
Incorporating mT5-large/mT5-xl models, context size of 2560, and constrained
decoding with depth-independent tags resulted in an increase of 3.4 percent
points. Furthermore, employing a 3-model ensemble provides another 1.2 percent
points raise. In the post-competition phase, we also evaluated an 8-model
ensemble, which delivered a final modest improvement of 0.3 percent points
and reached our best performance of 75.2\%.

\looseness-1
All these submissions choose the best model checkpoints for every corpus
independently. However, for deployment, a single checkpoint is more appropriate
-- therefore, we also assessed the single best-performing mT5-large checkpoint,
resulting in a 72.9\% score (0.8 percent points lower than choosing the best
mT5-large/mT5-xl checkpoint per corpus). The single best-performing mT5-xl
checkpoint achieved very similar performance of 73.0\%. We note that these
single-checkpoint submissions would comfortably win the shared task too.

\begin{table*}[t]
  \centering\setlength{\tabcolsep}{2.9pt}\scriptsize
\catcode`@ = 13\def@{\bfseries}
\catcode`! = 13\def!{\itshape}
\begin{tabular}{lcrrrrrrrrrrrrrrrrrr}
  \toprule
    Configuration & \makecell[r]{\llap{La}bel \\ \llap{Smooth}ing} & Avg &
    \texttt{ca} &
    \makecell[c]{\texttt{cs} \\ \texttt{\kern-.2em pcedt\kern-.2em}} &
    \makecell[c]{\texttt{cs} \\ \texttt{pdt}} &
    \makecell[c]{\texttt{de} \\ \texttt{parc}} &
    \makecell[c]{\texttt{de} \\ \texttt{pots}} &
    \makecell[c]{\texttt{en} \\ \texttt{gum}} &
    \makecell[c]{\texttt{en} \\ \texttt{parc}} &
    \texttt{es} &
    \texttt{fr} &
    \makecell[c]{\texttt{hu} \\ \texttt{\kern-.2em korko\kern-.2em}} &
    \makecell[c]{\texttt{hu} \\ \texttt{\kern-.2em szege\kern-.2em}} &
    \texttt{lt} &
    \makecell[c]{\texttt{no} \\ \texttt{\kern-.2em bookm\kern-.2em}} &
    \makecell[c]{\texttt{no} \\ \texttt{\kern-.2em nynor\kern-.2em}} &
    \texttt{pl} &
    \texttt{ru} &
    \texttt{tr} \\
  \midrule
  \noalign{\vskip 1pt}\multicolumn{20}{l}{\textsc{A) Constraing Decoding with Varying Depth and Depth-independent Tags}} \\[3pt]
    Depth 10 & !0.2 & @74.8 & @81.6 & 80.3 & @79.0 & 69.7 & 75.4 & @76.8 & 66.0 & 82.8 & @70.3 & @69.5 & 69.8 & 77.9 & 81.5 & 81.7 & 77.1 & 75.2 & 57.2 \\
    Depth 3 & !0.2 & \textcolor{red!1.3!black}{+0.0} & \textcolor{red!1.3!black}{--\kern 0.04em 0.1} & \textcolor{blue!100.0!black}{@+0.0} & \textcolor{red!1.4!black}{--\kern 0.04em 0.1} & \textcolor{red!0.0!black}{+0.0} & \textcolor{red!0.0!black}{+0.0} & \textcolor{red!0.9!black}{+0.0} & \textcolor{blue!0.0!black}{+0.0} & \textcolor{red!3.4!black}{--\kern 0.04em 0.3} & \textcolor{red!0.0!black}{@+0.0} & \textcolor{red!0.0!black}{@+0.0} & \textcolor{red!0.0!black}{+0.0} & \textcolor{blue!0.0!black}{+0.0} & \textcolor{red!4.7!black}{--\kern 0.04em 0.1} & \textcolor{blue!100.0!black}{@+0.0} & \textcolor{red!2.9!black}{+0.0} & \textcolor{red!0.0!black}{+0.0} & \textcolor{blue!100.0!black}{@+0.0} \\
\bottomrule
\end{tabular}

  \caption{Ablation experiments evaluated on the development sets (CoNLL score in \%) using the mT5-large model with context size 2560. We report the average of best 5 out of 7 runs, using for every corpus the single epoch achieving the highest average 5-run score.}
  \label{tab:decoding_ablations}
\end{table*}

\section{Ablations on the Development Set}
\label{sec:ablations}

To evaluate the effect of various hyperparameters, we perform further
experiments on the development set. Because we observed a significant
variance with different random seeds and we also observed divergence in some
training runs, we devised the following procedure to obtain credible results:
For each configuration, we perform 7 training runs and keep only the 5
ones with the best overall performance. We then want to perform early stopping
for every corpus. However, choosing for every corpus a different epoch in
every run could lead to maximization bias in case the results oscillate
considerably -- therefore, for every corpus, we choose the single epoch
achieving the highest average 5-run score (i.e., we use this epoch for all
5 runs). Finally, we either average or ensemble the 5 runs for every corpus.

\subsection{Pretrained Models and Context Sizes}
\label{sec:plm_and_context}

The effect of increasing context sizes on the mT5-large pretrained model
is presented in Table~\ref{tab:models_ablations}.A. The performance
improves consistently with increasing context size up to 2560; however,
context size 4096 deteriorates the performance slightly. Considering
context size 512, decreasing the context size by 128 to 384 decreases
the performance by 1.6 percent points, while increasing the context size by 128
to 768 increases it by 1.2 percent points, with performance improving up to
2 percent points for context length 2560.

For the mT5-xl pretrained model, the behavior is virtually analogous, as
captured by Table~\ref{tab:models_ablations}.B.

In Table~\ref{tab:models_ablations}.C, we compare the performance of different
pretrained models using the context size 512. We include different sizes of the
mT5 model~\citep{xue-etal-2021-mt5}, together with RemBERT~\citep{chung2021rethinking},
XLM-R base, and XLM-R large \citep{conneau-etal-2020-unsupervised}.%
\footnote{We do not include other base-sized models like
XLM-V~\citep{liang23XLMV} or mDeBERTaV3~\citep{he2023debertav3}, because they
lack behind the large-sized models.}

As expected, the increasingly bigger mT5 models improve the performance.
Somewhat surprisingly, the XLM-R-base surpasses mT5-base and XLM-R-large
and RemBERT surpass mT5-large. However, we discovered that the difference is
caused primarily by different tokenization: The mT5 tokenizer produces
on average more subwords than the XLM-R and RemBERT tokenizers, which
effectively decreases the context size of the mT5 models -- but the performance
is considerably dependent on the context size.

To expose the issue, Table~\ref{tab:models_ablations}.D compares various
pretrained models with different context sizes. Most importantly, we include
the performance of the XLM-R and RemBERT models using a context that would
be tokenized into 512 subwords by the mT5 tokenizer (presented in italics
and denoted by the \textit{mT5-512} context size). In these cases, the
performance is quite similar to the performance of the corresponding
mT5 model (with the notable exception of RemBERT's performance on Turkish,
which is considerably worse). However, the mT5 models support larger
context sizes (due to relative positional embeddings); already with context
size 768, the mT5 models surpass all models of corresponding size and context
size 512, ultimately providing the best results.

\subsection{Mention Decoding Algorithms}
\label{sec:decoding_ablations}

\begin{table*}[t]
  \centering\setlength{\tabcolsep}{2.3pt}\scriptsize
  \renewcommand\arraystretch{0.98}
\catcode`@ = 13\def@{\bfseries}
\catcode`! = 13\def!{\itshape}
\begin{tabular}{lrrrrrrrrrrrrrrrrrr}
  \toprule
    Configuration & Avg &
    \texttt{ca} &
    \makecell[c]{\texttt{cs} \\ \texttt{\kern-.2em pcedt\kern-.2em}} &
    \makecell[c]{\texttt{cs} \\ \texttt{pdt}} &
    \makecell[c]{\texttt{de} \\ \texttt{parc}} &
    \makecell[c]{\texttt{de} \\ \texttt{pots}} &
    \makecell[c]{\texttt{en} \\ \texttt{gum}} &
    \makecell[c]{\texttt{en} \\ \texttt{parc}} &
    \texttt{es} &
    \texttt{fr} &
    \makecell[c]{\texttt{hu} \\ \texttt{\kern-.2em korko\kern-.2em}} &
    \makecell[c]{\texttt{hu} \\ \texttt{\kern-.2em szege\kern-.2em}} &
    \texttt{lt} &
    \makecell[c]{\texttt{no} \\ \texttt{\kern-.2em bookm\kern-.2em}} &
    \makecell[c]{\texttt{no} \\ \texttt{\kern-.2em nynor\kern-.2em}} &
    \texttt{pl} &
    \texttt{ru} &
    \texttt{tr} \\
  \midrule
    Single Multilingual Model & @74.8 & @81.6 & @80.3 & @79.0 & @69.7 & @75.4 & @76.8 & @66.0 & @82.8 & @70.3 & @69.5 & @69.8 & @77.9 & @81.5 & @81.7 & @77.1 & @75.2 & @57.2 \\
    Per-Corpus Models & \textcolor{red!28.2!black}{--\kern 0.04em 3.7} & \textcolor{red!29.4!black}{--\kern 0.04em 1.4} & \textcolor{red!1.9!black}{--\kern 0.04em 0.5} & \textcolor{red!2.6!black}{--\kern 0.04em 0.4} & \textcolor{red!56.3!black}{--\kern 0.04em 7.7} & \textcolor{red!31.7!black}{--\kern 0.04em 3.3} & \textcolor{red!11.1!black}{--\kern 0.04em 1.6} & \textcolor{red!55.5!black}{--\kern 0.04em 7.6} & \textcolor{red!76.6!black}{--\kern 0.04em 1.5} & \textcolor{red!36.2!black}{--\kern 0.04em 2.0} & \textcolor{red!60.3!black}{--\kern 0.04em 9.1} & \textcolor{red!6.6!black}{--\kern 0.04em 1.0} & \textcolor{red!13.1!black}{--\kern 0.04em 3.0} & \textcolor{red!16.1!black}{--\kern 0.04em 2.3} & \textcolor{red!16.2!black}{--\kern 0.04em 2.9} & \textcolor{red!5.7!black}{--\kern 0.04em 1.0} & \textcolor{red!13.4!black}{--\kern 0.04em 2.0} & \textcolor{red!100.0!black}{--\kern 0.04em 15.8} \\
    Joint Czech Model &  &  & \textcolor{red!0.4!black}{--\kern 0.04em 0.1} & \textcolor{red!1.9!black}{--\kern 0.04em 0.3} &  &  &  &  &  &  &  &  &  &  &  &  &  &  \\
    Joint German Model &  &  &  &  & \textcolor{red!35.0!black}{--\kern 0.04em 4.8} & \textcolor{red!36.6!black}{--\kern 0.04em 3.9} &  &  &  &  &  &  &  &  &  &  &  &  \\
    Joint English Model &  &  &  &  &  &  & \textcolor{red!13.2!black}{--\kern 0.04em 1.9} & \textcolor{red!32.3!black}{--\kern 0.04em 4.5} &  &  &  &  &  &  &  &  &  &  \\
    Joint Parcorfull Model &  &  &  &  & \textcolor{red!32.6!black}{--\kern 0.04em 4.4} &  &  & \textcolor{red!18.0!black}{--\kern 0.04em 2.5} &  &  &  &  &  &  &  &  &  &  \\
    Joint Hungarian Model &  &  &  &  &  &  &  &  &  &  & \textcolor{red!39.1!black}{--\kern 0.04em 5.9} & \textcolor{red!6.8!black}{--\kern 0.04em 1.1} &  &  &  &  &  &  \\
    Joint Norwegian Model &  &  &  &  &  &  &  &  &  &  &  &  &  & \textcolor{red!8.8!black}{--\kern 0.04em 1.3} & \textcolor{red!9.7!black}{--\kern 0.04em 1.8} &  &  &  \\
    Zero-Shot Multilingual Models & \textcolor{red!100.0!black}{--\kern 0.04em 13.2} & \textcolor{red!100.0!black}{--\kern 0.04em 4.8} & \textcolor{red!100.0!black}{--\kern 0.04em 24.2} & \textcolor{red!100.0!black}{--\kern 0.04em 16.0} & \textcolor{red!100.0!black}{--\kern 0.04em 13.7} & \textcolor{red!100.0!black}{--\kern 0.04em 10.6} & \textcolor{red!100.0!black}{--\kern 0.04em 14.4} & \textcolor{red!100.0!black}{--\kern 0.04em 13.8} & \textcolor{red!100.0!black}{--\kern 0.04em 1.9} & \textcolor{red!100.0!black}{--\kern 0.04em 5.4} & \textcolor{red!100.0!black}{--\kern 0.04em 15.1} & \textcolor{red!100.0!black}{--\kern 0.04em 15.0} & \textcolor{red!100.0!black}{--\kern 0.04em 23.4} & \textcolor{red!100.0!black}{--\kern 0.04em 14.3} & \textcolor{red!100.0!black}{--\kern 0.04em 18.0} & \textcolor{red!100.0!black}{--\kern 0.04em 17.5} & \textcolor{red!100.0!black}{--\kern 0.04em 15.5} & \textcolor{red!5.1!black}{--\kern 0.04em 0.8} \\
  \bottomrule
\end{tabular}

  \caption{Ablation experiments evaluated on the development sets (CoNLL score in \%) using the mT5-large model with context size 2560. We report the average of best 5 out of 7 runs, using for every corpus the single epoch achieving the highest average 5-run score.}
  \label{tab:languages_ablations}
\end{table*}

\begin{table*}[t]
  \centering\setlength{\tabcolsep}{4.4pt}\scriptsize
  \renewcommand\arraystretch{0.98}
\catcode`@ = 13\def@{\bfseries}
\catcode`! = 13\def!{\itshape}
\begin{tabular}{lrrrrrrrrrrrrrrrrrr}
  \toprule
    Configuration & Avg &
    \texttt{ca} &
    \makecell[c]{\texttt{cs} \\ \texttt{\kern-.2em pcedt\kern-.2em}} &
    \makecell[c]{\texttt{cs} \\ \texttt{pdt}} &
    \makecell[c]{\texttt{de} \\ \texttt{parc}} &
    \makecell[c]{\texttt{de} \\ \texttt{pots}} &
    \makecell[c]{\texttt{en} \\ \texttt{gum}} &
    \makecell[c]{\texttt{en} \\ \texttt{parc}} &
    \texttt{es} &
    \texttt{fr} &
    \makecell[c]{\texttt{hu} \\ \texttt{\kern-.2em korko\kern-.2em}} &
    \makecell[c]{\texttt{hu} \\ \texttt{\kern-.2em szege\kern-.2em}} &
    \texttt{lt} &
    \makecell[c]{\texttt{no} \\ \texttt{\kern-.2em bookm\kern-.2em}} &
    \makecell[c]{\texttt{no} \\ \texttt{\kern-.2em nynor\kern-.2em}} &
    \texttt{pl} &
    \texttt{ru} &
    \texttt{tr} \\
  \midrule
  \noalign{\vskip 1pt}\multicolumn{19}{l}{\textsc{Mix Ratio Weights of Individual Corpora in Percents}} \\[3pt]
!Logarithmic & & !8.1 & !10.0 & !9.4 & !1.0 & !3.2 & !6.6 & !1.0 & !8.3 & !7.4 & !2.6 & !5.8 & !3.4 & !7.2 & !6.9 & !8.6 & !6.2 & !4.2 \\
!Uniform & & !5.9 & !5.9 & !5.9 & !5.9 & !5.9 & !5.9 & !5.9 & !5.9 & !5.9 & !5.9 & !5.9 & !5.9 & !5.9 & !5.9 & !5.9 & !5.9 & !5.9 \\
!Square Root & & !8.4 & !14.0 & !11.7 & !1.4 & !2.4 & !5.6 & !1.4 & !8.8 & !6.9 & !2.0 & !4.6 & !2.5 & !6.5 & !6.0 & !9.5 & !5.1 & !3.1 \\
!Linear & & !8.7 & !24.4 & !17.0 & !0.2 & !0.7 & !3.9 & !0.2 & !9.6 & !5.9 & !0.5 & !2.6 & !0.8 & !5.3 & !4.5 & !11.3 & !3.2 & !1.2 \\
  \midrule
  \noalign{\vskip 1pt}\multicolumn{19}{l}{\textsc{A) Average of 5 Runs Using for Every Corpus the Single Epoch Achieving the Highest Average 5-run Score}} \\[3pt]
    Logarithmic & 74.8 & 81.6 & 80.3 & 79.0 & 69.7 & 75.4 & @76.8 & 66.0 & 82.8 & 70.3 & 69.5 & @69.7 & 77.9 & 81.5 & 81.7 & 77.1 & 75.2 & @57.2 \\
    \quad w/o corpus id & \textcolor{red!39.9!black}{--\kern 0.04em 0.2} & \textcolor{blue!100.0!black}{@+0.2} & \textcolor{red!5.4!black}{--\kern 0.04em 0.1} & \textcolor{blue!17.2!black}{+0.1} & \textcolor{red!23.0!black}{--\kern 0.04em 0.4} & \textcolor{blue!4.9!black}{+0.1} & \textcolor{red!39.2!black}{--\kern 0.04em 0.3} & \textcolor{red!3.9!black}{--\kern 0.04em 0.2} & \textcolor{red!7.7!black}{+0.0} & \textcolor{blue!14.5!black}{+0.0} & \textcolor{red!16.2!black}{--\kern 0.04em 0.2} & \textcolor{red!36.7!black}{--\kern 0.04em 0.3} & \textcolor{blue!46.5!black}{+0.5} & \textcolor{blue!100.0!black}{@+0.2} & \textcolor{red!26.0!black}{--\kern 0.04em 0.4} & \textcolor{blue!100.0!black}{@+0.2} & \textcolor{blue!20.4!black}{+0.2} & \textcolor{red!80.4!black}{--\kern 0.04em 2.4} \\
    Uniform & \textcolor{red!74.3!black}{--\kern 0.04em 0.3} & \textcolor{red!24.5!black}{--\kern 0.04em 0.1} & \textcolor{red!100.0!black}{--\kern 0.04em 1.2} & \textcolor{red!100.0!black}{--\kern 0.04em 0.9} & \textcolor{blue!61.0!black}{+1.7} & \textcolor{red!1.7!black}{+0.0} & \textcolor{red!88.9!black}{--\kern 0.04em 0.8} & \textcolor{red!100.0!black}{--\kern 0.04em 4.2} & \textcolor{red!100.0!black}{--\kern 0.04em 0.3} & \textcolor{blue!30.8!black}{+0.1} & \textcolor{blue!100.0!black}{@+0.2} & \textcolor{red!49.4!black}{--\kern 0.04em 0.4} & \textcolor{blue!100.0!black}{@+1.0} & \textcolor{blue!29.8!black}{+0.0} & \textcolor{red!2.5!black}{--\kern 0.04em 0.1} & \textcolor{red!10.8!black}{+0.0} & \textcolor{red!69.6!black}{--\kern 0.04em 0.2} & \textcolor{red!4.5!black}{--\kern 0.04em 0.1} \\
    \quad w/o corpus id & \textcolor{red!100.0!black}{--\kern 0.04em 0.4} & \textcolor{red!100.0!black}{--\kern 0.04em 0.4} & \textcolor{red!57.0!black}{--\kern 0.04em 0.7} & \textcolor{red!63.3!black}{--\kern 0.04em 0.6} & \textcolor{blue!83.3!black}{+2.3} & \textcolor{blue!36.1!black}{+0.3} & \textcolor{red!100.0!black}{--\kern 0.04em 0.8} & \textcolor{blue!31.3!black}{+1.5} & \textcolor{red!52.3!black}{--\kern 0.04em 0.1} & \textcolor{red!100.0!black}{--\kern 0.04em 0.4} & \textcolor{red!100.0!black}{--\kern 0.04em 1.3} & \textcolor{red!62.5!black}{--\kern 0.04em 0.5} & \textcolor{red!100.0!black}{--\kern 0.04em 0.7} & \textcolor{red!100.0!black}{--\kern 0.04em 0.4} & \textcolor{red!100.0!black}{--\kern 0.04em 1.3} & \textcolor{red!100.0!black}{--\kern 0.04em 0.5} & \textcolor{red!100.0!black}{--\kern 0.04em 0.2} & \textcolor{red!100.0!black}{--\kern 0.04em 3.0} \\
    Square Root & \textcolor{red!12.2!black}{+0.0} & \textcolor{blue!80.2!black}{+0.2} & \textcolor{blue!60.9!black}{+0.5} & \textcolor{blue!49.2!black}{+0.4} & \textcolor{red!11.6!black}{--\kern 0.04em 0.2} & \textcolor{blue!100.0!black}{@+0.9} & \textcolor{red!78.7!black}{--\kern 0.04em 0.6} & \textcolor{red!49.2!black}{--\kern 0.04em 2.1} & \textcolor{red!47.7!black}{--\kern 0.04em 0.1} & \textcolor{blue!19.4!black}{+0.1} & \textcolor{red!55.3!black}{--\kern 0.04em 0.7} & \textcolor{red!16.3!black}{--\kern 0.04em 0.1} & \textcolor{blue!81.5!black}{+0.8} & \textcolor{blue!89.4!black}{+0.1} & \textcolor{red!12.9!black}{--\kern 0.04em 0.2} & \textcolor{blue!88.0!black}{+0.2} & \textcolor{blue!84.4!black}{+0.9} & \textcolor{red!21.9!black}{--\kern 0.04em 0.7} \\
    \quad w/o corpus id & \textcolor{blue!34.3!black}{+0.2} & \textcolor{blue!45.0!black}{+0.1} & \textcolor{blue!48.4!black}{+0.4} & \textcolor{blue!38.1!black}{+0.3} & \textcolor{blue!100.0!black}{@+2.7} & \textcolor{red!66.5!black}{--\kern 0.04em 0.9} & \textcolor{red!38.7!black}{--\kern 0.04em 0.3} & \textcolor{blue!22.0!black}{+1.1} & \textcolor{blue!22.4!black}{+0.1} & \textcolor{blue!15.9!black}{+0.0} & \textcolor{red!35.5!black}{--\kern 0.04em 0.4} & \textcolor{red!28.2!black}{--\kern 0.04em 0.2} & \textcolor{blue!8.8!black}{+0.1} & \textcolor{blue!66.0!black}{+0.1} & \textcolor{red!7.9!black}{--\kern 0.04em 0.1} & \textcolor{blue!52.0!black}{+0.1} & \textcolor{blue!41.3!black}{+0.5} & \textcolor{red!24.0!black}{--\kern 0.04em 0.7} \\
    Linear & \textcolor{blue!100.0!black}{@+0.4} & \textcolor{blue!58.6!black}{+0.1} & \textcolor{blue!100.0!black}{@+0.8} & \textcolor{blue!100.0!black}{@+0.7} & \textcolor{blue!22.2!black}{+0.6} & \textcolor{red!10.4!black}{--\kern 0.04em 0.1} & \textcolor{red!33.2!black}{--\kern 0.04em 0.2} & \textcolor{blue!100.0!black}{@+4.8} & \textcolor{blue!100.0!black}{@+0.3} & \textcolor{blue!100.0!black}{@+0.4} & \textcolor{red!68.0!black}{--\kern 0.04em 0.9} & \textcolor{red!46.4!black}{--\kern 0.04em 0.4} & \textcolor{blue!56.8!black}{+0.6} & \textcolor{red!68.2!black}{--\kern 0.04em 0.3} & \textcolor{blue!100.0!black}{@+0.1} & \textcolor{blue!72.0!black}{+0.2} & \textcolor{blue!100.0!black}{@+1.1} & \textcolor{red!11.1!black}{--\kern 0.04em 0.3} \\
    \quad w/o corpus id & \textcolor{blue!4.5!black}{+0.0} & \textcolor{blue!2.7!black}{+0.0} & \textcolor{blue!91.4!black}{+0.7} & \textcolor{blue!86.7!black}{+0.6} & \textcolor{red!100.0!black}{--\kern 0.04em 2.0} & \textcolor{red!100.0!black}{--\kern 0.04em 1.4} & \textcolor{red!89.8!black}{--\kern 0.04em 0.8} & \textcolor{blue!82.8!black}{+4.0} & \textcolor{blue!90.2!black}{+0.3} & \textcolor{red!6.9!black}{--\kern 0.04em 0.1} & \textcolor{red!35.1!black}{--\kern 0.04em 0.4} & \textcolor{red!100.0!black}{--\kern 0.04em 0.9} & \textcolor{blue!36.6!black}{+0.4} & \textcolor{blue!83.0!black}{+0.1} & \textcolor{red!7.1!black}{--\kern 0.04em 0.1} & \textcolor{blue!69.0!black}{+0.2} & \textcolor{blue!65.0!black}{+0.7} & \textcolor{red!26.7!black}{--\kern 0.04em 0.8} \\
  \midrule
  \noalign{\vskip 1pt}\multicolumn{19}{l}{\textsc{B) Average of 5 Runs Using for Every Run the Single Epoch Achieving the Highest Score Across All Corpora}} \\[3pt]
    Logarithmic & 74.8 & 81.7 & 79.9 & 78.6 & 71.5 & @76.2 & @76.6 & 67.9 & 82.8 & 70.4 & 68.3 & 69.4 & 78.0 & 81.4 & 81.5 & 76.9 & 74.6 & 55.5 \\
    \quad w/o corpus id & \textcolor{red!32.2!black}{--\kern 0.04em 0.2} & \textcolor{blue!19.5!black}{+0.0} & \textcolor{blue!11.6!black}{+0.1} & \textcolor{blue!13.6!black}{+0.2} & \textcolor{red!78.2!black}{--\kern 0.04em 1.9} & \textcolor{red!11.6!black}{--\kern 0.04em 0.3} & \textcolor{red!37.0!black}{--\kern 0.04em 0.3} & \textcolor{red!13.1!black}{--\kern 0.04em 0.9} & \textcolor{red!44.1!black}{--\kern 0.04em 0.2} & \textcolor{red!67.4!black}{--\kern 0.04em 0.4} & \textcolor{red!6.4!black}{+0.0} & \textcolor{red!41.1!black}{--\kern 0.04em 0.2} & \textcolor{red!19.5!black}{--\kern 0.04em 0.2} & \textcolor{blue!22.1!black}{+0.1} & \textcolor{red!13.6!black}{--\kern 0.04em 0.2} & \textcolor{blue!74.9!black}{+0.3} & \textcolor{blue!60.7!black}{+1.0} & \textcolor{red!12.0!black}{--\kern 0.04em 0.3} \\
    Uniform & \textcolor{red!100.0!black}{--\kern 0.04em 0.6} & \textcolor{red!48.6!black}{--\kern 0.04em 0.4} & \textcolor{red!100.0!black}{--\kern 0.04em 1.1} & \textcolor{red!100.0!black}{--\kern 0.04em 0.9} & \textcolor{blue!5.4!black}{+0.1} & \textcolor{red!37.5!black}{--\kern 0.04em 1.0} & \textcolor{red!100.0!black}{--\kern 0.04em 0.8} & \textcolor{red!100.0!black}{--\kern 0.04em 6.7} & \textcolor{red!100.0!black}{--\kern 0.04em 0.4} & \textcolor{red!24.8!black}{--\kern 0.04em 0.2} & \textcolor{blue!100.0!black}{@+1.0} & \textcolor{blue!100.0!black}{@+0.1} & \textcolor{red!18.1!black}{--\kern 0.04em 0.2} & \textcolor{red!16.3!black}{--\kern 0.04em 0.1} & \textcolor{blue!57.3!black}{+0.2} & \textcolor{red!13.1!black}{--\kern 0.04em 0.1} & \textcolor{blue!28.3!black}{+0.5} & \textcolor{blue!0.2!black}{+0.0} \\
    \quad w/o corpus id & \textcolor{red!94.6!black}{--\kern 0.04em 0.6} & \textcolor{red!100.0!black}{--\kern 0.04em 0.7} & \textcolor{red!50.4!black}{--\kern 0.04em 0.6} & \textcolor{red!57.1!black}{--\kern 0.04em 0.5} & \textcolor{blue!72.8!black}{+1.0} & \textcolor{red!61.2!black}{--\kern 0.04em 1.6} & \textcolor{red!63.7!black}{--\kern 0.04em 0.5} & \textcolor{red!9.9!black}{--\kern 0.04em 0.6} & \textcolor{red!39.3!black}{--\kern 0.04em 0.1} & \textcolor{red!100.0!black}{--\kern 0.04em 0.6} & \textcolor{blue!30.9!black}{+0.3} & \textcolor{red!100.0!black}{--\kern 0.04em 0.5} & \textcolor{red!100.0!black}{--\kern 0.04em 0.9} & \textcolor{red!24.4!black}{--\kern 0.04em 0.1} & \textcolor{red!100.0!black}{--\kern 0.04em 1.3} & \textcolor{red!100.0!black}{--\kern 0.04em 0.5} & \textcolor{blue!49.1!black}{+0.8} & \textcolor{red!100.0!black}{--\kern 0.04em 3.0} \\
    Square Root & \textcolor{red!24.1!black}{--\kern 0.04em 0.2} & \textcolor{red!6.1!black}{--\kern 0.04em 0.1} & \textcolor{blue!71.9!black}{+0.8} & \textcolor{blue!58.4!black}{+0.7} & \textcolor{red!100.0!black}{--\kern 0.04em 2.5} & \textcolor{red!7.3!black}{--\kern 0.04em 0.2} & \textcolor{red!6.7!black}{--\kern 0.04em 0.1} & \textcolor{red!62.8!black}{--\kern 0.04em 4.2} & \textcolor{red!34.1!black}{--\kern 0.04em 0.1} & \textcolor{blue!11.9!black}{+0.0} & \textcolor{blue!98.5!black}{+0.9} & \textcolor{red!83.0!black}{--\kern 0.04em 0.4} & \textcolor{blue!42.5!black}{+0.2} & \textcolor{blue!74.0!black}{+0.3} & \textcolor{blue!9.4!black}{+0.0} & \textcolor{blue!100.0!black}{@+0.4} & \textcolor{blue!90.3!black}{+1.5} & \textcolor{blue!33.4!black}{+0.4} \\
    \quad w/o corpus id & \textcolor{blue!34.4!black}{+0.1} & \textcolor{red!22.2!black}{--\kern 0.04em 0.2} & \textcolor{blue!48.9!black}{+0.6} & \textcolor{blue!46.5!black}{+0.6} & \textcolor{blue!100.0!black}{@+1.3} & \textcolor{red!83.4!black}{--\kern 0.04em 2.1} & \textcolor{red!21.2!black}{--\kern 0.04em 0.2} & \textcolor{red!11.3!black}{--\kern 0.04em 0.7} & \textcolor{blue!34.4!black}{+0.2} & \textcolor{blue!100.0!black}{@+0.1} & \textcolor{red!2.3!black}{+0.0} & \textcolor{red!71.6!black}{--\kern 0.04em 0.4} & \textcolor{red!14.2!black}{--\kern 0.04em 0.1} & \textcolor{blue!53.8!black}{+0.2} & \textcolor{blue!34.9!black}{+0.1} & \textcolor{blue!35.8!black}{+0.1} & \textcolor{blue!72.0!black}{+1.2} & \textcolor{blue!100.0!black}{@+1.1} \\
    Linear & \textcolor{blue!100.0!black}{@+0.3} & \textcolor{blue!100.0!black}{@+0.2} & \textcolor{blue!100.0!black}{@+1.1} & \textcolor{blue!100.0!black}{@+1.1} & \textcolor{red!30.7!black}{--\kern 0.04em 0.7} & \textcolor{red!76.2!black}{--\kern 0.04em 1.9} & \textcolor{red!27.6!black}{--\kern 0.04em 0.2} & \textcolor{blue!100.0!black}{@+3.8} & \textcolor{blue!100.0!black}{@+0.5} & \textcolor{red!1.9!black}{--\kern 0.04em 0.1} & \textcolor{red!100.0!black}{--\kern 0.04em 0.7} & \textcolor{red!29.4!black}{--\kern 0.04em 0.1} & \textcolor{blue!56.4!black}{+0.3} & \textcolor{red!100.0!black}{--\kern 0.04em 0.4} & \textcolor{blue!100.0!black}{@+0.3} & \textcolor{blue!20.5!black}{+0.1} & \textcolor{blue!100.0!black}{@+1.6} & \textcolor{red!2.0!black}{+0.0} \\
    \quad w/o corpus id & \textcolor{blue!39.4!black}{+0.1} & \textcolor{blue!8.9!black}{+0.0} & \textcolor{blue!90.5!black}{+1.0} & \textcolor{blue!87.5!black}{+1.0} & \textcolor{red!85.8!black}{--\kern 0.04em 2.1} & \textcolor{red!100.0!black}{--\kern 0.04em 2.5} & \textcolor{red!20.7!black}{--\kern 0.04em 0.2} & \textcolor{blue!34.4!black}{+1.3} & \textcolor{blue!50.7!black}{+0.2} & \textcolor{red!10.4!black}{--\kern 0.04em 0.1} & \textcolor{blue!36.3!black}{+0.4} & \textcolor{red!89.4!black}{--\kern 0.04em 0.5} & \textcolor{blue!100.0!black}{@+0.5} & \textcolor{blue!100.0!black}{@+0.4} & \textcolor{blue!83.3!black}{+0.3} & \textcolor{blue!95.3!black}{+0.4} & \textcolor{blue!59.2!black}{+1.0} & \textcolor{blue!69.7!black}{+0.8} \\
  \bottomrule
\end{tabular}

  \caption{Ablation experiments evaluated on the development sets (CoNLL score in \%) using the mT5-large model with context size 2560. We report the average of best 5 out of 7 runs.}
  \label{tab:mix_ratios_ablations}
\end{table*}

\begin{table*}[t]
  \centering\setlength{\tabcolsep}{3.6pt}\scriptsize
\catcode`@ = 13\def@{\bfseries}
\catcode`! = 13\def!{\itshape}
\begin{tabular}{lrrrrrrrrrrrrrrrrrr}
  \toprule
    Configuration & Avg &
    \texttt{ca} &
    \makecell[c]{\texttt{cs} \\ \texttt{\kern-.2em pcedt\kern-.2em}} &
    \makecell[c]{\texttt{cs} \\ \texttt{pdt}} &
    \makecell[c]{\texttt{de} \\ \texttt{parc}} &
    \makecell[c]{\texttt{de} \\ \texttt{pots}} &
    \makecell[c]{\texttt{en} \\ \texttt{gum}} &
    \makecell[c]{\texttt{en} \\ \texttt{parc}} &
    \texttt{es} &
    \texttt{fr} &
    \makecell[c]{\texttt{hu} \\ \texttt{\kern-.2em korko\kern-.2em}} &
    \makecell[c]{\texttt{hu} \\ \texttt{\kern-.2em szege\kern-.2em}} &
    \texttt{lt} &
    \makecell[c]{\texttt{no} \\ \texttt{\kern-.2em bookm\kern-.2em}} &
    \makecell[c]{\texttt{no} \\ \texttt{\kern-.2em nynor\kern-.2em}} &
    \texttt{pl} &
    \texttt{ru} &
    \texttt{tr} \\
  \midrule
  \noalign{\vskip 1pt}\multicolumn{19}{l}{\textsc{A) Ensembles for the mT5-large Model for Various Context Sizes}} \\[3pt]
    Average of 5 runs, 512 & 72.8 & 78.1 & 78.1 & 76.9 & 70.7 & 75.4 & 75.6 & @67.4 & 80.3 & 68.6 & 70.6 & 67.3 & 77.4 & 77.8 & 78.7 & 75.8 & 71.1 & 48.6 \\
    Ensemble of 5 runs, 512 & \textcolor{blue!30.5!black}{+1.0} & \textcolor{blue!17.6!black}{+0.8} & \textcolor{blue!27.6!black}{+0.8} & \textcolor{blue!25.0!black}{+0.7} & \textcolor{blue!100.0!black}{@+3.1} & \textcolor{blue!100.0!black}{@+1.3} & \textcolor{blue!41.2!black}{+0.5} & \textcolor{red!30.7!black}{--\kern 0.04em 0.4} & \textcolor{blue!22.4!black}{+0.8} & \textcolor{blue!24.8!black}{+0.6} & \textcolor{blue!100.0!black}{@+1.2} & \textcolor{blue!19.7!black}{+0.7} & \textcolor{blue!92.8!black}{+1.6} & \textcolor{blue!18.1!black}{+0.9} & \textcolor{blue!24.3!black}{+0.9} & \textcolor{blue!43.1!black}{+1.0} & \textcolor{blue!24.6!black}{+1.5} & \textcolor{blue!7.2!black}{+0.8} \\
    Average of 5 runs, 768 & \textcolor{blue!36.5!black}{+1.2} & \textcolor{blue!57.4!black}{+2.5} & \textcolor{blue!39.9!black}{+1.2} & \textcolor{blue!50.5!black}{+1.5} & \textcolor{red!72.0!black}{--\kern 0.04em 0.7} & \textcolor{red!0.0!black}{+0.0} & \textcolor{blue!65.8!black}{+0.9} & \textcolor{red!100.0!black}{--\kern 0.04em 1.4} & \textcolor{blue!42.5!black}{+1.5} & \textcolor{blue!52.4!black}{+1.3} & \textcolor{red!53.8!black}{--\kern 0.04em 0.6} & \textcolor{blue!60.3!black}{+2.1} & \textcolor{blue!23.5!black}{+0.4} & \textcolor{blue!57.3!black}{+2.7} & \textcolor{blue!62.7!black}{+2.2} & \textcolor{blue!18.5!black}{+0.4} & \textcolor{blue!43.4!black}{+2.7} & \textcolor{blue!29.1!black}{+3.3} \\
    Average of 5 runs, 2560 & \textcolor{blue!61.2!black}{+2.0} & \textcolor{blue!81.1!black}{+3.5} & \textcolor{blue!73.7!black}{+2.2} & \textcolor{blue!70.5!black}{+2.1} & \textcolor{red!100.0!black}{--\kern 0.04em 1.0} & \textcolor{red!0.0!black}{+0.0} & \textcolor{blue!96.1!black}{+1.2} & \textcolor{red!99.6!black}{--\kern 0.04em 1.4} & \textcolor{blue!69.8!black}{+2.5} & \textcolor{blue!67.5!black}{+1.7} & \textcolor{red!100.0!black}{--\kern 0.04em 1.1} & \textcolor{blue!70.8!black}{+2.5} & \textcolor{blue!32.4!black}{+0.5} & \textcolor{blue!79.1!black}{+3.7} & \textcolor{blue!85.6!black}{+3.0} & \textcolor{blue!57.6!black}{+1.3} & \textcolor{blue!66.1!black}{+4.1} & \textcolor{blue!75.2!black}{+8.6} \\
    Ensemble of 5 runs, 2560 & \textcolor{blue!100.0!black}{@+3.3} & \textcolor{blue!100.0!black}{@+4.3} & \textcolor{blue!100.0!black}{@+3.0} & \textcolor{blue!100.0!black}{@+3.0} & \textcolor{blue!73.8!black}{+2.3} & \textcolor{blue!100.0!black}{@+1.3} & \textcolor{blue!100.0!black}{@+1.3} & \textcolor{red!57.2!black}{--\kern 0.04em 0.8} & \textcolor{blue!100.0!black}{@+3.6} & \textcolor{blue!100.0!black}{@+2.5} & \textcolor{blue!88.3!black}{+1.1} & \textcolor{blue!100.0!black}{@+3.5} & \textcolor{blue!100.0!black}{@+1.8} & \textcolor{blue!100.0!black}{@+4.6} & \textcolor{blue!100.0!black}{@+3.5} & \textcolor{blue!100.0!black}{@+2.3} & \textcolor{blue!100.0!black}{@+6.3} & \textcolor{blue!100.0!black}{@+11.5} \\
  \midrule
  \noalign{\vskip 1pt}\multicolumn{19}{l}{\textsc{B) Ensembles for the mT5-xl Model for Various Context Sizes}} \\[3pt]
    Average of 5 runs, 512 & 73.3 & 77.5 & 78.4 & 77.2 & 73.9 & 76.1 & 75.4 & 72.9 & 80.1 & 68.4 & 70.3 & 67.2 & 77.2 & 77.7 & 78.3 & 76.1 & 71.3 & 47.6 \\
    Ensemble of 5 runs, 512 & \textcolor{blue!23.8!black}{+0.8} & \textcolor{blue!22.6!black}{+1.1} & \textcolor{blue!25.0!black}{+0.9} & \textcolor{blue!22.2!black}{+0.8} & \textcolor{red!51.8!black}{--\kern 0.04em 2.3} & \textcolor{blue!100.0!black}{@+0.2} & \textcolor{blue!34.2!black}{+0.8} & \textcolor{blue!100.0!black}{@+1.9} & \textcolor{blue!31.4!black}{+1.1} & \textcolor{blue!31.5!black}{+1.1} & \textcolor{blue!73.1!black}{+0.9} & \textcolor{blue!44.0!black}{+1.8} & \textcolor{blue!53.8!black}{+1.6} & \textcolor{blue!25.4!black}{+1.1} & \textcolor{blue!15.2!black}{+0.8} & \textcolor{blue!40.2!black}{+1.0} & \textcolor{blue!17.8!black}{+1.3} & \textcolor{blue!4.0!black}{+0.3} \\
    Average of 5 runs, 768 & \textcolor{blue!31.1!black}{+1.1} & \textcolor{blue!46.1!black}{+2.2} & \textcolor{blue!36.6!black}{+1.3} & \textcolor{blue!45.6!black}{+1.7} & \textcolor{red!100.0!black}{--\kern 0.04em 4.4} & \textcolor{blue!76.7!black}{+0.1} & \textcolor{blue!53.9!black}{+1.3} & \textcolor{blue!47.0!black}{+0.9} & \textcolor{blue!48.0!black}{+1.7} & \textcolor{blue!46.5!black}{+1.5} & \textcolor{red!84.0!black}{--\kern 0.04em 1.3} & \textcolor{blue!47.7!black}{+1.9} & \textcolor{blue!49.4!black}{+1.5} & \textcolor{blue!62.1!black}{+2.6} & \textcolor{blue!43.3!black}{+2.2} & \textcolor{blue!20.6!black}{+0.5} & \textcolor{blue!35.9!black}{+2.6} & \textcolor{blue!30.6!black}{+2.4} \\
    Average of 5 runs, 2560 & \textcolor{blue!55.7!black}{+1.9} & \textcolor{blue!70.8!black}{+3.4} & \textcolor{blue!71.5!black}{+2.6} & \textcolor{blue!70.0!black}{+2.6} & \textcolor{red!98.5!black}{--\kern 0.04em 4.4} & \textcolor{blue!76.7!black}{+0.1} & \textcolor{blue!72.4!black}{+1.7} & \textcolor{blue!49.9!black}{+1.0} & \textcolor{blue!77.7!black}{+2.8} & \textcolor{blue!61.9!black}{+2.0} & \textcolor{red!100.0!black}{--\kern 0.04em 1.5} & \textcolor{blue!54.1!black}{+2.2} & \textcolor{blue!39.1!black}{+1.2} & \textcolor{blue!88.6!black}{+3.7} & \textcolor{blue!71.2!black}{+3.6} & \textcolor{blue!55.5!black}{+1.4} & \textcolor{blue!74.5!black}{+5.3} & \textcolor{blue!74.6!black}{+5.7} \\
    Ensemble of 5 runs, 2560 & \textcolor{blue!100.0!black}{@+3.5} & \textcolor{blue!100.0!black}{@+4.9} & \textcolor{blue!100.0!black}{@+3.6} & \textcolor{blue!100.0!black}{@+3.7} & \textcolor{blue!100.0!black}{@+2.4} & \textcolor{blue!100.0!black}{@+0.2} & \textcolor{blue!100.0!black}{@+2.3} & \textcolor{blue!58.9!black}{+1.1} & \textcolor{blue!100.0!black}{@+3.6} & \textcolor{blue!100.0!black}{@+3.3} & \textcolor{blue!100.0!black}{@+1.3} & \textcolor{blue!100.0!black}{@+4.0} & \textcolor{blue!100.0!black}{@+3.0} & \textcolor{blue!100.0!black}{@+4.1} & \textcolor{blue!100.0!black}{@+5.0} & \textcolor{blue!100.0!black}{@+2.5} & \textcolor{blue!100.0!black}{@+7.1} & \textcolor{blue!100.0!black}{@+7.6} \\
  \bottomrule
\end{tabular}

  \caption{Ablation experiments evaluated on the development sets (CoNLL score in \%). We report the average/ensemble of best 5 out of 7 runs, using for every corpus the single epoch achieving the highest average score.}
  \label{tab:ensembles_ablations}
\end{table*}

The effects of the mention decoding algorithm and label smoothing
are elaborated in Table~\ref{tab:decoding_ablations}. First, label
smoothing has very little effect on the results.

When predicting mentions via depth-independent tags, the maximum possible
number of opened multi-word mentions (\textit{depth}) must be specified. The effect of
using depths 1, 2, 3, and 10 is presented in
Table~\ref{tab:decoding_ablations}.A. While the maximum depth in the training
data is 12, the performance of using depth 10 and 3 is virtually unchanged;
only depth 2 and depth 1 deteriorate performance. If the speed of the decoding
is an issue, using depth~3 provides the fastest decoder without decreasing
performance.

The difference between using depth-independent and depth-dependent tags
during constrained decoding is quantified in
Table~\ref{tab:decoding_ablations}.B -- depth-independent tags provide
a minor improvement of 0.3 percent points. When greedy decoding is used
instead of constrained decoding, the performance drops by one percent point.

Using conditional random fields for mention decoding provides marginally
worse performance compared to using constrained decoding with depth-independent
tags. Furthermore, explicitly disallowing invalid transitions (by assigning them
transition weight $-\infty$ in the transition weight matrix manually) has
virtually no effect, demonstrating that the CRF decoder has learned the
transition weights successfully.

\subsection{The Effect Of Multilingual Data}

In Table~\ref{tab:languages_ablations}, we analyze the effect of using
various combinations of corpora during training.

Compared to using all corpora for single-model training, relying
solely on the training data of a given corpus deteriorates the
performance dramatically by 3.7 percent points on average. The decrease
is smallest for the largest corpora (Czech and Polish ones).

Concatenating all corpora of a given language (and both ParCorFull corpora that
are translations of each other; we utilized uniform mix ratios) generally
improves the performance compared to using the individual corpora, but does not
reach the performance of using all corpora together.

\subsection{Zero-shot Multilingual Evaluation}

When training without the corpus ids, the model is able to perform prediction
on unknown languages. Leveraging this observation, we perform zero-shot
evaluation by training multilingual models on corpora from all but one
language and then evaluating the performance on the omitted-language corpora.
The results are displayed on the last line of
Table~\ref{tab:languages_ablations}.

Overall, the results are significantly worse by 13.2 percent points.
However, such performance is most likely better than the performance
of the baseline system of~\citet{prazak-etal-2021-multilingual}, which has
17.9 less percent points on the test set than CorPipe.

Turkish demonstrates the smallest decrease in the zero-shot evaluation, even
when it uses an alphabet with several unique characters. On the other hand, the
small decrease in the performance of Catalan, Spanish, and French can be
explained by similarities among these languages.

\subsection{Mix Ratios of the Multilingual Data}
\label{sec:mix_ratios}

Next, we compare the effect of various mix ratios during all-corpora training.

We consider \textit{logarithmic}, \textit{uniform}, \textit{square root},
and \textit{linear} mix ratios described in
Section~\ref{sec:multilingual_training_data}. First, their values normalized to
percentages are presented in the first part of Table~\ref{tab:mix_ratios_ablations}.

\looseness-1
We then evaluate the effect of using a specific mix ratio and either utilizing
or omitting the corpus ids during training in
Table~\ref{tab:mix_ratios_ablations}.A. In accordance with findings in
\citet{straka-strakova-2022-ufal}, the corpus ids have no deterministic effect,
and the mix ratios influence the system performance surprisingly little (with
\textit{uniform} being the worst, \textit{logarithmic} and \textit{square root}
very similar and better, and \textit{linear} the best). When considering
the largest corpora (especially Czech, Polish, and Spanish), their performance
improves with increasing mix ratios, presumably because of underfitting with
small mix ratios; however, the effect on other corpora is mixed.

\looseness-1
The evaluation methodology allows each corpus to use a checkpoint from
a different epoch of the training. Therefore, it could be possible that different
mixing ratios influence the best epochs of individual corpora and that
with some mixing ratios, the best epochs are more homogeneous. On that account,
Table~\ref{tab:mix_ratios_ablations}.B performs the evaluation differently -- for
each of the 5 runs, we choose the epoch with the best overall performance on all
corpora, and employ the checkpoint from this epoch for all corpora; different
runs can utilize different epochs. Nevertheless, the results are very much
similar.

\subsection{Ensembling}

The effect of ensembling the 5 runs (instead of averaging them) is captured
in Table~\ref{tab:ensembles_ablations}. For the context size 512, the ensemble
delivers an additional 1 percent point with the mT5-large pretrained model
and 0.8 percent points with the mT5-xl model. For the context size 2560, the
improvement is even slightly larger, 1.3 and 1.6 percent points for the
mT5-large and mT5-xl models, respectively.

\section{Conclusions}

We presented the winning entry to the \CRAClong~\cite{sharedtask-findings}.
The system is an improved version of our earlier multilingual coreference
pipeline CorPipe~\cite{straka-strakova-2022-ufal}, and it surpasses other
participants by a large margin of 4.5 percent points. When ensembling is
not desired, we also offer a single multilingual checkpoint for all 17
corpora surpassing other submissions by 2.6 percent points.
The source code is available at {\small\url{https://github.com/ufal/crac2023-corpipe}}.

\section*{Acknowledgements}

This work has been supported by the Grant Agency of the Czech Republic,
project EXPRO LUSyD (GX20-16819X), and has been using data provided by
the LINDAT/CLARIAH-CZ Research Infrastructure
({\footnotesize\url{https://lindat.cz}}) of the Ministry of Education,
Youth and Sports of the Czech Republic (Project No. LM2023062).

\section*{Limitations}
The presented system has demonstrated its performance only on a limited set of
12 languages, and heavily depends on a large pretrained model, transitively
receiving its limitations and biases.

Furthermore, the practical applicability on plain text inputs depends also on
empty node prediction, whose performance has not yet been evaluated.

Training with the mT5-large pretrained model requires a 40GB GPU, which we
consider affordable; however, training with the mT5-xl pretrained model needs
nearly four times as much GPU memory.

\bibliography{anthology,custom}
\bibliographystyle{acl_natbib}

\end{document}